\documentclass[conference]{IEEEtran}
\usepackage{times}
\usepackage[T1]{fontenc}

\usepackage[numbers]{natbib}
\usepackage{multicol}
\usepackage[bookmarks=true]{hyperref}

\usepackage{xcolor}
\usepackage{pbox}
\usepackage{graphicx}
\usepackage{subcaption}
\usepackage{float}
\usepackage{array}
\usepackage{xcolor}
\usepackage{algorithm} 
\usepackage{algorithmic}  
\usepackage[algo2e]{algorithm2e} 

\newcommand{\ignorethis}[1]{}
\usepackage{stackengine}
\newcommand\xrowht[2][0]{\addstackgap[.5\dimexpr#2\relax]{\vphantom{#1}}}
\usepackage{amsthm,amsmath,amssymb,amsfonts}
\usepackage{cleveref}
\newtheorem{theorem}{Theorem}[section]

\newtheorem{definition}[theorem]{Definition}
\newtheorem{remark}[theorem]{Remark}

\DeclareMathOperator*{\argmax}{argmax}

\newcommand\blfootnote[1]{%
  \begingroup
  \renewcommand\thefootnote{}\footnote{#1}%
  \addtocounter{footnote}{-1}%
  \endgroup
}
\usepackage[
top = 0.75in,
bottom = 0.8in,
left   = 0.75in,
right  = 0.75in]{geometry}

\begin{document}

% paper title
\title{\vspace*{0.25in} Rapidly Adaptable Legged Robots\\ via Evolutionary Meta-Learning}
%%CF.2.24: "Fast Adaptation *to* Dynamics Changes" would be better English. Here are a few other paper title ideas:
% Adaptable Legged Robots via Evolutionary Meta-Learning
% Fast Adaptation for Legged Robots via Evolutionary Meta-Learning
% Rapidly Adaptable Robots via Evolutionary Meta-Learning
%% KC: Evolutionary Meta-Learning Enables Fast Adaptation for Legged Robots.

% You will get a Paper-ID when submitting a pdf file to the conference system
%\author{Author Names Omitted for Anonymous Review.}

\author{\authorblockN{Xingyou Song$^{\ast}$\authorrefmark{2} 
Yuxiang Yang$^{\ast}$\authorrefmark{2}\authorrefmark{4} 
Krzysztof Choromanski\authorrefmark{2} \\
Ken Caluwaerts\authorrefmark{2}
Wenbo Gao\authorrefmark{3} 
Chelsea Finn\authorrefmark{2} 
Jie Tan\authorrefmark{2}}
\authorblockA{\authorrefmark{2}Robotics at Google \authorrefmark{3}Columbia University}}

\maketitle

\blfootnote{$^\ast$Equal contribution.}
\blfootnote{$\mathsection$Work performed during the Google AI Residency Program. \textcolor{blue}{\url{http://g.co/airesidency}}}
\blfootnote{Video can be found at \textcolor{blue}{\url{https://youtu.be/_QPMCDdFC3E}}.} 
\blfootnote{Code can be found at \textcolor{blue}{\url{https://github.com/google-research/google-research/tree/master/es_maml}}.} 
\blfootnote{Corresponding Google AI Blog post can be found at \textcolor{blue}{\url{https://ai.googleblog.com/2020/04/exploring-evolutionary-meta-learning-in.html}}.}

\begin{abstract}
Learning adaptable policies is crucial for robots to operate autonomously in our complex and quickly changing world. In this work, we present a new meta-learning method that allows robots to quickly adapt to changes in dynamics. In contrast to gradient-based meta-learning algorithms that rely on second-order gradient estimation, we introduce a more noise-tolerant Batch Hill-Climbing adaptation operator and combine it with meta-learning based on evolutionary strategies. Our method significantly improves adaptation to changes in dynamics in high noise settings, which are common in robotics applications. We validate our approach on a quadruped robot that learns to walk while subject to changes in dynamics. We observe that our method significantly outperforms prior gradient-based approaches, enabling the robot to adapt its policy to changes based on less than 3 minutes of real data.
\end{abstract}

\IEEEpeerreviewmaketitle

\section{Introduction} \label{sec:introduction}
Deep reinforcement learning (RL) holds the promise of automatically acquiring locomotion controllers for legged robots~\cite{haarnoja2018learning, hwangbo2019learning, sim-to-real, xie2019iterative}. Typically, these methods train a control policy in a computationally efficient simulation environment and then deploy the learned policy to hardware (i.e. sim-to-real). However, the policies learned with deep RL often only work well in environments highly similar to those they were trained on. Any mismatch or non-trivial changes to the environment can require re-training from scratch. For robots to operate autonomously in our complex and ever-changing world, it is crucial that they can adapt their control policies quickly to accommodate these changes. For example, a legged robot may need to change its locomotion gait if its battery level is low, if a motor is broken, or if it is carrying a heavy load.

\textit{Meta-learning} leverages previous experience to explicitly train a policy to be adaptable, by training across many different tasks or environment conditions. One such approach specifically designed to make policies adaptable is \textit{model agnostic meta-learning} (MAML)~\cite{PG-MAML}; MAML performs adaptation via gradient updates on policy parameters using a small amount of data, optimizing so that such updates are fast and efficient. Since then, numerous works have improved upon this paradigm~\cite{ES-MAML, norml}, usually using simulators such as MuJoCo~\cite{mujoco} and PyBullet~\cite{pybullet} for evaluating and comparing algorithms.

However, applying these techniques to real robots is challenging due to the noise present in real world environments. Even with the same control policy, the trajectories of different runs can quickly diverge and lead to significantly different reward signals. This stochasticity renders gradient estimates unreliable during adaptation, unless we dramatically increase the amount of data collected. Different initial conditions, hardware wear-and-tear, observation noise, and action noise make exact repetition of trials difficult. This problem is especially evident for legged locomotion for which contact events are frequent and a small difference in contacts can generate significantly different trajectories. 

\begin{figure}[t]
\centering
\includegraphics[width=1.0\linewidth]{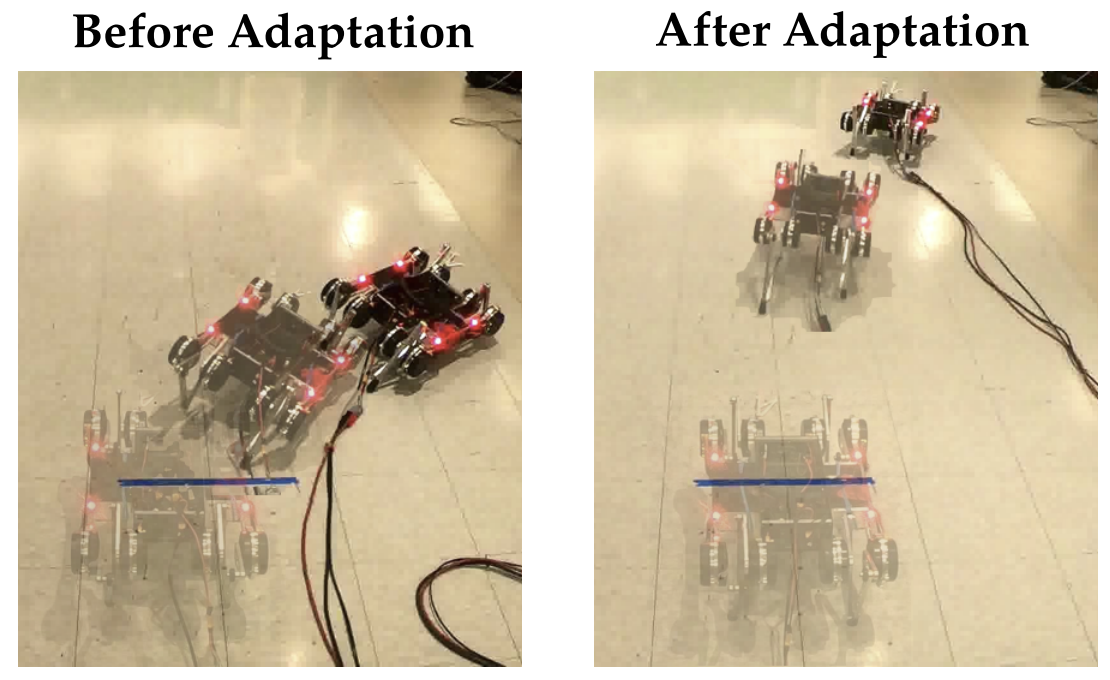}
\caption{Our algorithm quickly adapts a legged robot's policy to dynamics changes. In this example, the battery voltage dropped from 16.8V to 10V which reduced motor power, and a 500g mass was also placed on the robot's side, causing it to turn rather than walk straight. Based on only 50 episodes (150s) of real world data, our method adapted the policy to significantly improve performance.}
\label{fig:real_robot_snapshot}
\end{figure}

The primary motivation for this work is: How we can ensure effective adaptation in \emph{highly noisy environments}? When using the class of policy gradient algorithms \cite{REINFORCE, PPO, TRPO} which require an action \textit{distribution}, we may at first glance wish to \textit{decrease} the entropy of this distribution as the high-noise problem might be exacerbated by the additional randomness from the policy's actions. However, we may also wish to \textit{increase} this entropy as well, as the policy needs random actions for \textit{exploration} to determine the type of environment to adapt to. These two competing objectives, which seek to both decrease and increase action entropy, may cause complications when using policy gradients, as noted in \cite{DPG,DDPG}. 

The class of evolutionary strategies (ES) \cite{salimans, ARS} algorithms are attractive approaches to rectify these problems. ES leverages parameter-space search, which allows it to learn deterministic policies that are particularly conducive to adaptation. In this work, we combine ES with meta-learning, adopting the ES-MAML framework \cite{ES-MAML}. This framework allows us to explore a more powerful and non-differential adaptation process. To combat stochasticity in noisy real-world environments, we introduce a noise-tolerant and data-efficient adaptation operator based on Batch Hill-Climbing (BHC). Since this operator is not differentiable, it is difficult to integrate with the state-of-the-art policy-gradient-based MAML paradigms (PG-MAML) \cite{PG-MAML}. Hence, we use ES-MAML, and show that ES-MAML with BHC is not only empirically more effective than MAML, but also theoretically justifiable under conditions with strong noise.

Furthermore, ES-MAML allows adaptation using compact linear policies, which have been shown to qualitatively produce more stable behaviors \cite{ARS, simplicity}. While policy gradients are also able to train linear policies \cite{simplicity}, effective adaptation with PG-MAML relies on flexible representational capacity \cite{maml-and-universality}; thus, linear policies and meta-learning may not combine well for policy gradients.

We validate our approach on the Minitaur, a quadruped robot platform. We demonstrate that a policy trained purely in simulation using our method can not only overcome the sim-to-real gap, but also adapts successfully to two difficult new real-world environments: 1) a low battery level with extra payload and 2) a slippery walking surface. We further compare our method with PG-MAML approaches and conduct ablation studies on various design choices of our method. We show that our method adapts faster and achieves high rewards post-adaptation, even in environments that are out of distribution from training. We also show that our BHC method is more noise tolerant than other Hill-Climbing methods, which is essential for real-world experiments. In summary, the contributions of this paper include:

\begin{enumerate}
\item We introduce and integrate BHC, a non-differentiable, highly-efficient and noise-tolerant adaptation operator, into ES-MAML to learn adaptable locomotion policies for legged robots. 
\item We successfully demonstrate fast adaptation on a quadruped robot. It requires only 50 episodes totalling 150 seconds of robot data, to adapt the control policies in two drastically different environments.
\item We provide a thorough evaluation of ES-MAML with BHC both in simulation and on a real robot, which demonstrates its advantage over gradient-based meta-learning techniques.
\end{enumerate}

\section{Related Work}

%%CF.2.24: Here are some related meta-learning works that are quite relevant and should be cited:
% Nagabandi et al. Learning to Adapt in Dynamic Real-World Environments (ICLR) -- model-based meta-RL with MAML, results on real legged robots for dynamics adaptation <-- Done (xsong)
% Houthooft et al. Evolved Policy Gradients - model-free meta-RL method that uses evolutionary methods in the outer loop
% Meta-Learning by the Baldwin Effect -- another evolutionary meta-learning method <-- Done (xsong)
% more meta-RL papers: RL^2, Wang et al. Learning to Reinforcement Learn, SNAIL, ProMP, MAESN (Abhishek Gupta et al), VariBAD (Zintgraf et al) <-- Done (xsong)
%%CF.2.24: There are also a lot more sim2real works that would be worth citing. Igor Mordatch has a paper on sim2real for the Darwin robot, for example. Also Fereshteh's paper on CAD2RL, which was basically the original domain randomization paper <-- Done (xsong)
\subsection{Learning Locomotion}
Reinforcement learning is becoming a popular method to develop locomotion controllers for legged robots \cite{tedrake2005learning, tianyubiped, xie2019iterative, singla2019realizing, icra04, AAAI04, sim-to-real, hwangbo2019learning}. Learning can be performed in simulation and followed by a sim-to-real transfer phase \cite{sim-to-real, hwangbo2019learning} or carried out directly on real hardware \cite{haarnoja2018learning, ha2018automated}. Since legged robots can walk in diverse environments, the ability to change their learned policies to fit new environments is important. Prior work has investigated online system identification \cite{preparing-for-unknown}, action transformation \cite{hanna2017grounded}, precomputed behavior-performance map \cite{cully2015robots}, strategy optimization \cite{yu2018policy, meta-strategy-optimization} and meta-learning \cite{learning_to_adapt} for quick adaptation of locomotion controllers.

\subsection{Policy Transfer}
Transferring a learned policy to different dynamics has been a long-standing problem in reinforcement learning. \textit{Domain Randomization (DR)} trains the policy with experiences collected from multiple dynamics, so that the learned policy can remain robust across a number of environments~\cite{sim-to-real, peng2018sim, cad2rl}. More recently, advanced ways of domain randomization have been proposed such as hierarchical domain randomization~\cite{hierarchical-sim-to-real}, curriculum sampling~\cite{active-dr}, and inverse dynamics \cite{transfer_inverse}. Although DR may work effectively for sim-to-real transfer, it trades optimality for robustness and makes the learning problem much harder.

Another class of approaches augment the state space with a latent vector, which represents the environment changes. The latent vector can be based on actual physics parameters~\cite{preparing-for-unknown, yu2018policy} or based on a learned embedding of the trajectory~\cite{rakelly2019efficient}. During adaptation, the algorithm either infers the latent representation of the environment through system identification~\cite{preparing-for-unknown}, or directly optimizes the latent variable for optimal performance~\cite{meta-strategy-optimization, yu2018policy}. While this approach has demonstrated adaptation in several real-robot scenarios~\cite{yu2018policy, meta-strategy-optimization}, choosing the parameters of the latent representation requires careful tuning.

\subsection{Meta-Learning}
More relevant to our work, meta-learning~\cite{PG-MAML} has been used for policy transfer in reinforcement learning where the problem was defined as finding a \textit{meta-policy} that can be adapted to a new task in a few gradient steps. The original Model-Agnostic Meta-Learning (MAML) algorithm~\cite{PG-MAML} uses policy gradient methods as the policy adaptation operator, which we refer to as \textit{PG-MAML}. Several improvements of PG-MAML have been proposed: stochastic gradient adaptation~\cite{finn}, improving estimation of the Hessian~\cite{taming, promp}, or replacing second derivatives with multiple gradient steps ($\mathrm{Reptile}$, \cite{nichol}). Adaptation can be performed explicitly on a latent space \cite{MRLSES, arndt2019meta, varibad}, or learned via RNN-based methods~\cite{RL2, L2RL, SNAIL}. However, most of these methods are so far only tested in basic simulated environments. 

In order to leverage MAML for dynamics adaptation in real world environments, \cite{learning_to_adapt} combines \textit{model-based methods} with MAML, and presents results on a small hexpod robot. Arndt et al.~\cite{arndt2019meta} is one of the few works to date that has applied the model-free PG-MAML to train a robotic arm to hit a hockey puck. In contrast, our locomotion environment is significantly noisier \cite{sim-to-real} due to frequent and discrete contact events inherent to locomotion, as well as high uncertainties about the robot's dynamics due to cheap hardware. 

\section{Preliminaries} \label{subsec:preliminaries}
\subsection{Problem Setup}
We formulate the meta reinforcement learning problem as an optimization over a distribution of tasks: $T_i\in P(\mathcal{T})$, where each task is a Markov Decision Process (MDP). In our setting, all tasks share the same state space $\mathcal{S}$, the same action space $\mathcal{A}$, and most importantly \textit{the same reward function} $r:\mathcal{S}\times\mathcal{A}\to\mathbb{R}$. However, given a fixed state $s_{t}$, each task can have a different transition distribution $p(s_{t+1}|s_{t}, a_{t})$ which corresponds to different dynamics. We parameterize a policy $\pi:\mathcal{S}\to\mathcal{A}$ as a neural network with parameters $\theta\in\Theta$, denoted as $\pi_\theta$.

We seek to find a policy that can adapt to different tasks efficiently. Formally, given an adaptation procedure $U:\Theta\times\mathcal{T}\to\Theta$, we optimize for a meta-policy $\pi_{\theta_{meta}}$ that maximizes the expected return of the adapted policies:
\begin{equation}\label{eq:maml_objective}
\begin{aligned}
\theta_{meta} = \argmax_{\theta} \mathbb{E}_{T \sim \mathcal{P}(\mathcal{T})}
\left[f^{T}(U(\theta, T))\right]
\end{aligned}
\end{equation}
\text{where} $f^{T}(\theta) = \mathbb{E}_{(s_{0:H}, a_{1:H})\sim \pi_{\theta}, T}\left[\sum_{t=1}^H{r(s_t, a_t)}\right]$ is the expected total sum of rewards using policy $\pi_{\theta}$ over the dynamics corresponding to task $T$. In the MAML algorithm~\cite{finn}, the update operator $U$ corresponds to one or a few steps of gradient descent, and maximization of the objective in Eq.~\ref{eq:maml_objective} is performed with gradient-based optimization. We define the \textit{adaptation gap} between the meta-policy $\theta_{meta}$ and the adapted policy $U(\theta_{meta}, T)$ as the difference $f^{T}(U(\theta_{meta}, T)) - f^{T}(\theta_{meta})$, or also its expectation across $\mathcal{P}(\mathcal{T})$ when multiple tasks are involved, depending on context.

\subsection{Meta-Learning with Evolutionary Strategies}
Previous works~\cite{baldwin, evolved_policy_gradients} replace some components of the MAML algorithm such as the outer-loop with evolutionary methods, but still rely on policy gradients to train differentiable stochastic policies. ES-MAML \cite{ES-MAML} is an algorithm under the meta-learning paradigm which uses techniques for optimizing \textit{blackbox} objectives. 

In our setting, we use a low variance \textit{antithetic} estimator~\cite{ARS} for the outer loop of the ES-MAML algorithm \cite{ES-MAML}. This estimates the gradient $\nabla_{\theta} \mathbb{E}_{\mathbf{g} \sim \mathcal{N}(0,\mathbf{I}) }[F(\theta + \sigma \mathbf{g})] = \frac{1}{2\sigma} \mathbb{E}[f(\theta + \sigma \mathbf{g}) - f(\theta - \sigma \mathbf{g})]$ of the Gaussian-smoothed version of the outer loop objective $F(\theta) = \mathbb{E}_{T \sim \mathcal{P}(\mathcal{T})}[ f^{T}(U(\theta, T))]$. We present this algorithm formally in Algorithm~\ref{algo:general_zo_sgd}.

\begin{figure}[ht]
\centering
\begin{minipage}{1.0\linewidth}
\begin{algorithm}[H]
\SetAlgoLined
\KwData{initialized meta-policy $\theta_{meta}$, meta step size $\beta$}
\While{\text{not done}}{
    Sample $n$ tasks $T_{1},\ldots,T_{n} \sim \mathcal{P}(\mathcal{T})$ and i.i.d. vectors $\mathbf{g}_{1},\ldots,\mathbf{g}_{n} \sim \mathcal{N}(0,\mathbf{I})$\;
    \ForEach{$(T_{i}, \mathbf{g}_{i})$}
    {
        $\>\> v_{i}^{+} \gets f^{T_{i}}(U(\theta_{meta} + \sigma \mathbf{g}_{i},T_{i}))$ \\
        $v_{i}^{-} \gets f^{T_{i}}(U(\theta_{meta} - \sigma \mathbf{g}_{i},T_{i}))$ \\
        $v_i \gets \frac{1}{2} (v_{i}^{+} - v_{i}^{-})$ \\
    }
    Update $\theta_{meta} \gets \theta_{meta} + \frac{\beta}{\sigma n} \sum_{i=1}^{n} v_{i} \mathbf{g}_{i}$
 }
\caption{ES-MAML (Antithetic Variant) using general adaptation operator $U(\cdot, T)$.}
\label{algo:general_zo_sgd}
\end{algorithm}
\end{minipage}
\end{figure}

We use the iterative Hill-Climbing (HC) operator $U_{HC}$ in \cite{ES-MAML}, which has been shown to be a strong adaptation operator for classic MAML benchmarks. HC has multiple benefits: for example in the deterministic case, it enforces monotonic improvement in the objective. More formally, given an adaptation rate $\alpha$ and a query number $(Q)$ for number of rollouts allowed during finetuning, the Hill-Climbing step is defined as simply $\theta^{(q+1)}=\argmax_{\theta \in \{\theta^{(q)}, \theta^{(q)} + \alpha\mathbf{g} \}}f(\theta)$, for which we apply iteratively $\theta_{meta} \rightarrow \theta^{(1)} \rightarrow ... \rightarrow \theta^{(Q)}$ to produce a Hill-Climbing operator $U_{HC}(\theta_{meta}) = \theta^{(Q)}$. Note that $U_{HC}$ is a non-differentiable operator which cannot be replicated in the PG case, as the Hill-Climbing operator uses $\argmax$. 

\section{Evolutionary Meta-Learning in Noisy Environments} \label{noisy_objectives}
%\subsection{Effective Adaptation in Noisy Environments} 
In Subsection~\ref{subsec:overview}, we first give a description of the noise model as well as an overview of our modifications to the HC operator. We then provide detailed analysis of these modifications and their comparisons in Subsection~\ref{subsec:adaptation_mechanisms}. In the following Section~\ref{sec:legged_robots}, we describe how this method can be used for training adaptable policies for legged robots.

\subsection{Overview} \label{subsec:overview}
Our goal is to apply evolutionary meta-learning on legged robots in the real world. One bottleneck that we must address in real world settings is the abundance of noise, which may affect the trajectory of the Hill-Climbing adaptation. As a reasonable and general abstraction, we choose to model the noise $\varepsilon$ \textit{additively} in our evaluation of $f$ as $\widetilde{f}(\theta, \varepsilon) = f(\theta) + \varepsilon$ and discuss its assumptions in Subsec. ~\ref{subsec:adaptation_mechanisms}. In Table \ref{table:hill_climbing}, we generalize the original (termed "Sequential") Hill-Climbing method from \cite{ES-MAML}. We modify the 1-step update rule $\theta^{(q)} \rightarrow \theta^{(q+1)}$ to allow multiple ($P>1$) parallel evaluations. We derived two variants, \textit{Average} and \textit{Batch}, for dealing with noisy evaluations. 

\begin{table}[ht]
\begin{center}
\scalebox{1.03}{
\begin{tabular}{|c|l|} 
\hline\xrowht[()]{10pt}
Sequential & $\theta^{(q+1)} = \argmax\limits_{\theta\in {\{\theta^{(q)}, \theta^{(q)} + \alpha\mathbf{g} \}}} f(\theta)$ \\
\hline\xrowht[()]{10pt}
Average & $\theta^{(q+1)} =  \argmax\limits_{\theta\in {\{\theta^{(q)}, \theta^{(q)} + \alpha\mathbf{g} \}}}\frac{1}{P}\sum_{i=1}^P \widetilde{f}(\theta, \varepsilon_i)$ \\
\hline\xrowht[()]{10pt}
Batch & $\theta^{(q+1)} = \argmax\limits_{ \theta\in \{\theta^{(q)}, \theta^{(q)} + \alpha\mathbf{g}_{1}, ... ,\theta^{(q)} + \alpha\mathbf{g}_{P} \}} \widetilde{f}(\theta, \varepsilon)  $ \\
\hline
\end{tabular}
}
\caption{We present the 3 Hill-Climbing steps for comparison: \textit{Sequential, Average,} and \textit{Batch}.}
\label{table:hill_climbing}
\end{center}
\end{table}

\subsection{Adaptation Mechanisms} \label{subsec:adaptation_mechanisms}
\textit{\textbf{Average}}: A straightforward way to handle noise is to average noisy evaluations. We use empirical averages $\widehat{f}(\theta) = \frac{1}{P} \sum_{i=1}^{P} \widetilde{f}(\theta, \varepsilon_{i})$ when evaluating a fixed parameter $\theta$. However, this method has two limitations: (1) Low sample efficiency and (2) Restrictive noise assumptions.

\textit{Low sample efficiency}. Real robot experiments are expensive. Assuming a fixed budget $(Q \cdot P)$ of \textit{total} objective evaluations allowed during the adaptation, this averaging method sacrifices exploration for accuracy by lowering $Q$ (the number of new parameters sampled) and increasing $P$ (the number of parallel objective evaluations). Since it is often infeasible to measure the noise level in the real world, it is difficult to find a right balance between $P$ and $Q$ in practice. 

\textit{Restrictive noise assumptions.} This method inherently assumes the existence of a true \textit{expected} objective $ \mathbb{E}_{\varepsilon \sim \mathcal{D}}[\widetilde{f}(\theta, \varepsilon)]$ which can be approximated by $\widehat{f}(\theta)$, under a strict assumption that errors $\varepsilon$ are i.i.d. sampled from a zero-mean distribution $\mathcal{D}$. The i.i.d. assumption can be unrealistic when experimenting in the real world where noise can easily be correlated (see Subsection~\ref{subsec:noise_adaptation} for a detailed description in the Minitaur case). The zero-mean assumption is also unrealistic, as \cite{RBO}, which also touches on the Minitaur case, assumes the objective can be corrupted with \textit{unbounded} magnitude \textit{adversarially}, under bounds of the frequency of such corruptions. The adversarial assumption, which we will also use, avoids overly specific and complex modelling of the error by covering a variety of noise types using general \textit{worst-case} theoretical guarantees. While \cite{RBO} approaches this problem by estimating a biased gradient using regression, we must use an alternative method, as the Hill-Climbing method inherently does not estimate gradients. This leads to the following simple, yet effective adaptation step:

\textit{\textbf{Batch}}:
In order to maintain sample efficiency and reduce strong assumptions on noise, we propose the "Batch" Hill-Climbing method by modifying the adaptation operator to sample a \textit{batch} of perturbations before performing the argmax, shown in Table~\ref{table:hill_climbing}. Our batch method still maintains a diverse population of $(Q \cdot P)$ new parameters. Note that it takes the argmax based on \textit{noisy} evaluations of the objective from $P$ candidate parameters $\theta^{(q)} + \alpha \mathbf{g}_{i}, \> \> \forall i \in \{1,2,...,P\}$. This exact variant (termed \textit{Gradientless Descent}) originates from \cite{GLD}, and was proven to possess convergence properties (for the convex/concave objective case) and is also independent to monotonic transformations of the reward (such as scaling) which can be useful in new test domains, when the model of the noise is generated by a bounded deterministic function $\varepsilon = h(\theta)$ of the input.

The noise in our real-robot setting exceeds even these assumptions, since the noise is non-deterministic and there can also be non-independent catastrophic failures. However, it turns out that Batch Hill-Climbing is robust even to this extreme level of noise. We assume that the reward estimates $f(\cdot)$ for all candidate permutations $\{ \theta^{(q)} + \alpha \mathbf{g}_1, \ldots, \theta^{(q)} + \alpha  \mathbf{g}_P\}$ may be corrupted with additive error $|\epsilon| \leq \Lambda$, and that up to $W$ rewards may be corrupted with unbounded negative errors. If the number of candidates $P$ is sufficiently large relative to $W$, then the average regret of the Batch Hill-Climbing algorithm can be bounded, up to irreducible error from $\Lambda$. That is, with high probability, we have $\frac{1}{Q}\sum_{q=0}^Q \left[f(\theta^{opt}) - f(\theta^{(q)}) \right] \leq \frac{c_1}{\sqrt{Q}} + c_2$, where $c_1, c_2$ are constants depending on the concavity of the reward $f$, the size of the domain, and the level of noise $\Lambda$. We prove this novel claim formally in the Appendix.

\section{Application to Legged Robots} \label{sec:legged_robots}
We use the Minitaur quadruped robot from Ghost Robotics \cite{ghostrobotics} as our hardware platform.The Minitaur is actuated by 8 direct-drive motors, two for each leg. The robot is equipped with an IMU sensor to measure the orientation and the angular velocity of the robot's base, and motor encoders to measure its joint angles. In addition, we attach motion capture markers on the base of the robot. During the adaptation phase, we use a motion capture system to track the position of the robot to calculate the reward signal. We tethered the robot with a workstation, which runs the control loop at 166Hz. To validate our algorithm on the real robot, we first train a meta-policy in simulation, and then adapt the policy to two real-world tasks where the dynamics have changed significantly.

\subsection{Training in Simulation} \label{subsec:task_setup}
Since training in the real world is costly, we train our meta-policy in simulation and rely on our method to adapt the learned policy to various real world environments. The observation space includes the roll and pitch angle of the robot base, as well as all 8 motor angles, and a phase variable \cite{li2019learning}. The action space consists of the desired swing and extension of each leg \cite{sim-to-real}. The reward function is
\begin{displaymath}
r(t)=\min(v, v_{\text{max}})dt - 0.005\sum_{i=1}^8\tau_i \omega_i dt
\end{displaymath} 
where $v,v_{\text{max}}$ is the current and maximum forward velocity of the robot, $\tau_i, \omega_i$ is the torque and angular velocity of each motor. The reward encourages the robot to walk at maximum speed, while penalizing energy expenditure. $v_{\text{max}}$ starts at 0 when $t=0$, increases linearly to 1.3m/s at 0.6s, and remains constant afterwards.

\begin{table}[b]
    \centering
    \scalebox{1.12}{
    \begin{tabular}{|c|c|c|}
        \hline
       \textbf{Parameter}  & \textbf{Lower Bound} & \textbf{Upper Bound} \\
       \hline\hline
        Base Mass & 75\% & 150\%\\\hline
        Leg Mass & 75\% & 150\%\\\hline
        Battery Voltage & 14.8V & 16.8V\\\hline
        Motor Viscous Damping & 0.0 & 0.02\\\hline
        Motor Strength & 70\% & 100\% \\\hline
        Contact Friction & 0.75 & 1.5\\\hline
        Control Latency & 0s & 0.05s \\\hline
    \end{tabular}
    }
    \caption{\label{table:parameter_range}Range of parameters randomized.}
\end{table}

We leverage a physics simulation of the robot using PyBullet \cite{pybullet}. During training, each task samples a physics parameter from Table \ref{table:parameter_range}. All tasks are trained with a maximum horizon of 500 steps (3 seconds). To prevent unsafe behaviors, we terminate an episode early if the center height drops below 13cm, or if the robot orientation deviates too far from the upright position. For ES-MAML, we use 300 workers for perturbations, and the same hyperparameters as found in~\cite{ES-MAML}. For PG-MAML, we use the same implementation as~\cite{PG-MAML} and tuned hyperparameters using gridsearch. For all experiments, our error bars are plotted with 1 standard deviation from the mean.

In order to pretrain our policy to effectively use a noise-reduction adaptation operator when applied to real-world tasks, we inject observation noise sampled from a unit Gaussian distribution in the simulated training environment, and also randomize the Minitaur's initialization.

\subsection{Adaptation Tasks in the Real World} \label{subsec:real_robot}
On the real robot, we test the adaptation of our method on two difficult tasks. Note that the changes between training and test environments are not only due to sim-to-real inaccuracies, but also include additional changes in battery level, center of mass, or foot friction.

\begin{figure}[t]
    \centering
    \includegraphics[width=\linewidth]{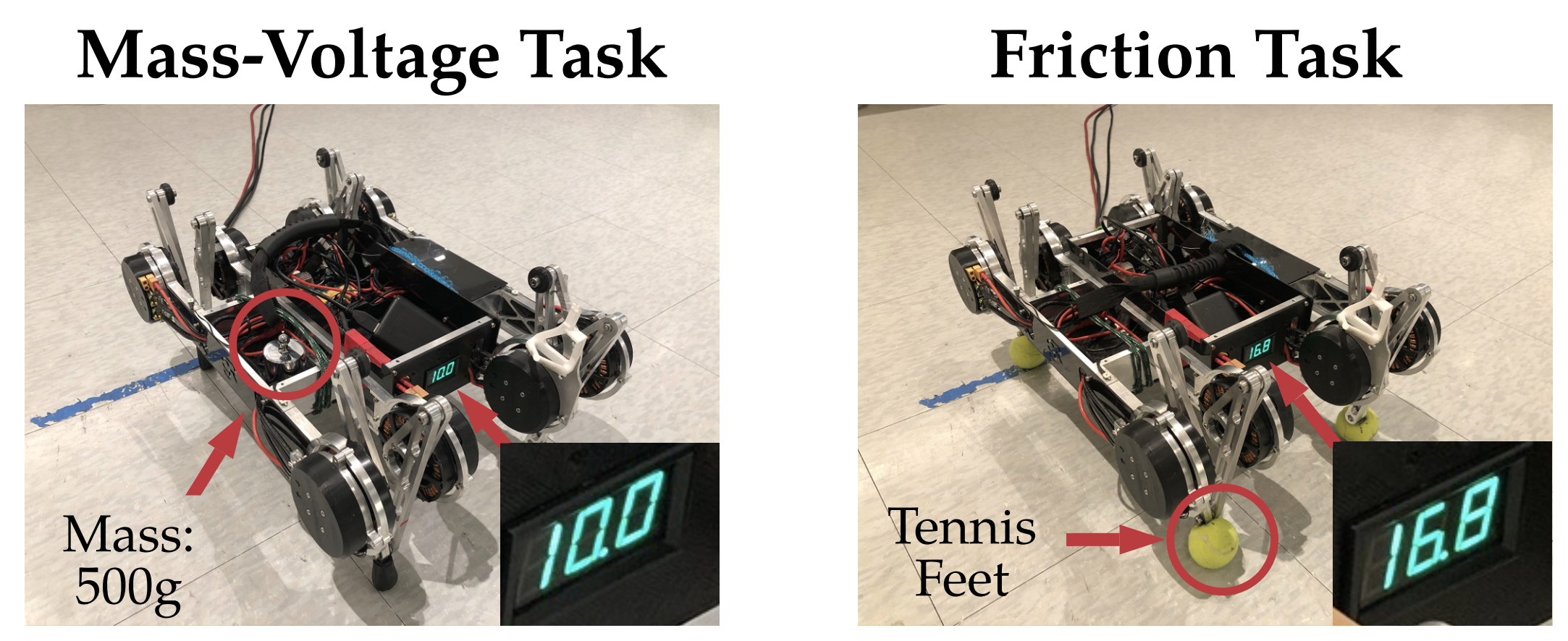}
    \caption{\label{fig:real_world_tasks} Illustration of two real robot tasks, each with a red circle covering the physical modification, as well as depicting the battery voltages used.}
\end{figure}

\textbf{Mass-Voltage Task} (Fig.~\ref{fig:real_world_tasks}, Left): In this task, we reduced the voltage of the power supply from 16.8V to 10V and added a mass of 500 grams on the right side of the robot (about 8\% of robot mass). Note that although we randomized the battery voltage and base mass during training, the test voltage (10V) is far outside the training range (14.8V-16.8V) and the position of center-of-mass is not randomized.

\textbf{Friction Task} (Fig.~\ref{fig:real_world_tasks}, Right): To create a slippery contact, we replaced all rubber feet of the Minitaur robot with tennis balls, which significantly reduced the friction coefficient of contact.

\subsection{Real Robot Results} \label{subsec:real_robot_results}
\begin{figure}[ht]
  \begin{subfigure}{\linewidth}
    \centering
    \includegraphics[keepaspectratio, width=\linewidth]{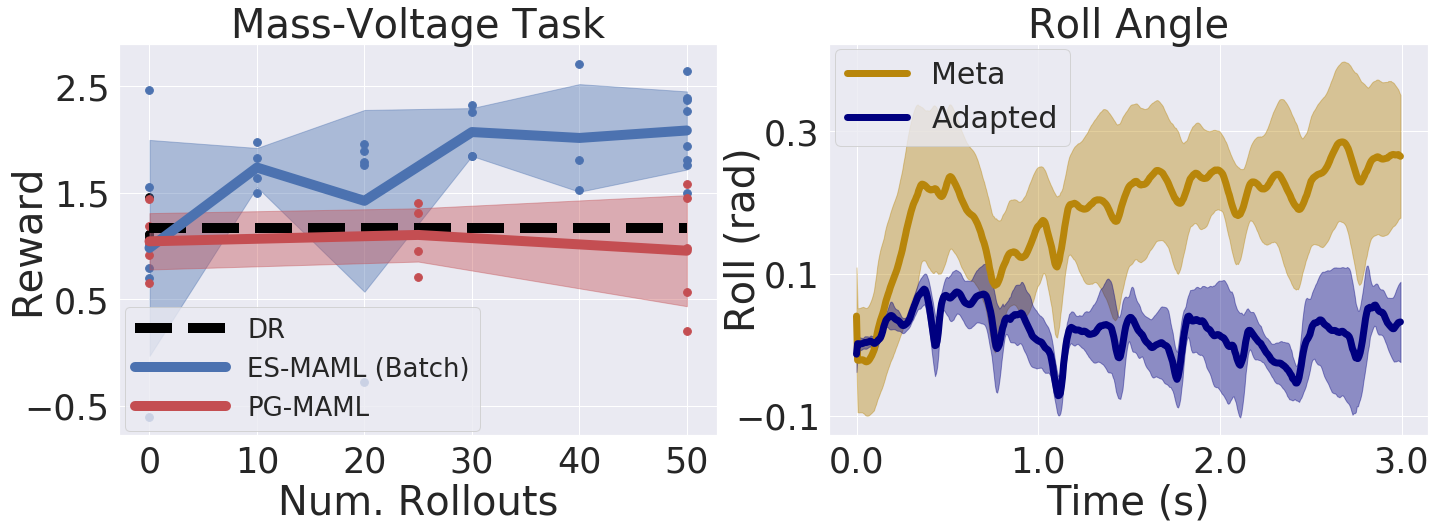}
    \caption{\label{fig:massvoltage} Real robot result for \textbf{Mass-Voltage Task}. (\textbf{Left}): ES-MAML increases its performance as we use more rollouts during adaptation, outperforming domain randomization (DR) and PG-MAML. (\textbf{Right}): Roll Angle (deviation of Minitaur body from z-axis, to measure turning) is displayed for each of meta-policy and adapted policy from ES-MAML. Adaptation stabilizes the roll angle to $\approx 0$, reducing turning.}
  \end{subfigure}
  \begin{subfigure}{\linewidth}
  \centering
    \includegraphics[keepaspectratio, width=\linewidth]{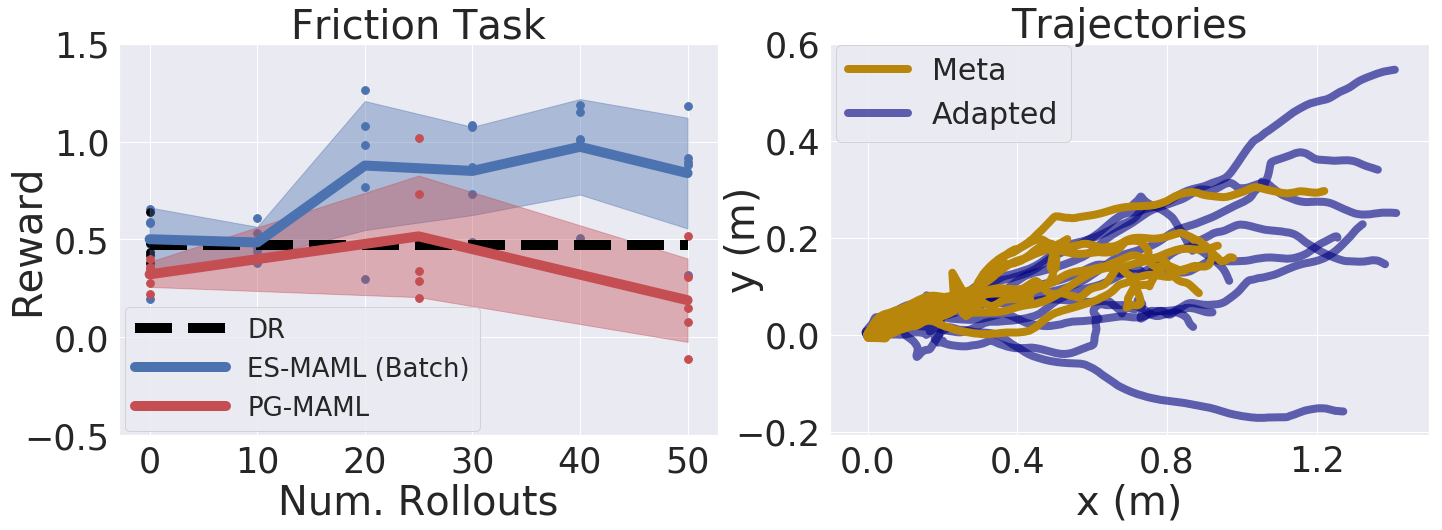}
    \caption{\label{fig:friction_curve} Real robot result for \textbf{Friction Task}. (\textbf{Left}): ES-MAML once again improves performance throughout Batch Hill-Climbing, while PG-MAML actually performs \textit{worse} after 2 gradient steps of adaptation. (\textbf{Right}): Sample rollouts are displayed from both the meta-policy and adapted policy from ES-MAML, and shows that adaptation allows longer trajectories. }
  \end{subfigure}
  \caption{\label{fig:friction} Real robot results showing raw rewards and metrics collected from rollouts using our method (Batch Hill-Climbing) from both the meta-policy and adapted policy. For each task, we allow 50 rollouts for adaptation and set Batch Hill-Climbing to use $(Q, P) = (5, 10)$. }
\end{figure}

For the Mass-Voltage Task, the added weight tilts the robot to the right, which together with the reduced motor voltage, turns the robot towards right when we ran the meta-policy on the robot. Our method adapts the policy effectively within 50 real-world rollouts (150 seconds of data), improving the return by $100\%$ (Fig.~\ref{fig:massvoltage}, Left). The adaptation corrects the tilt, and returns the robot to a balanced roll angle (Fig.~\ref{fig:massvoltage}, Right). As a result, the robot is able to walk farther and more straight as shown in Fig.~\ref{fig:real_robot_snapshot} from Section~\ref{sec:introduction}.

For the Friction Task, when applying the meta-policy, the extremely low friction causes the robot to slip constantly and leads to the chaotic trajectory distribution (Fig.~\ref{fig:friction}, Left). The fine-tuned policy is still able to significantly increase forward walking distance, highlighting the superior performance of our method in noisy environments.

As a comparison, we trained the policy with Domain Randomization (DR) using ES. When using our method, although the meta policy achieves a similar performance as DR, the adapted policy significantly outperforms the DR policy. We also attempted adaptation with PG-MAML on the real robot using the same number of 50 rollouts (2 gradient step and 25 rollouts for each). For both tasks, although the meta policy performs similarly to that of ES-MAML, the policy does not improve after adaptation (Fig.~\ref{fig:massvoltage}), and even performs worse in the Friction Task (Fig.~\ref{fig:friction_curve}). 

\section{Simulation Analysis and Discussion} \label{sec:ablation}
To further demonstrate the effectiveness of our method, we perform additional ablation studies over several key components of our method in simulation. We aim to answer the following questions in our analysis:
\begin{itemize}
    \item Does adaptation happen using our method? Qualitatively, how does the adaptation improve the performance?
    \item Can our method adapt more effectively and efficiently than PG-MAML?
    \item How does stochasticity in the environment affect training, and how does our Batch Hill-Climbing adaptation algorithm perform in noisy environments?
\end{itemize}

For a thorough evaluation, we performed a large amount of experiments in simulation to answer the above three questions. 

\subsection{Learned Adaptation Behaviors}
\begin{figure}[t]
  \begin{subfigure}[t]{\linewidth}
    \centering
    \includegraphics[keepaspectratio, width=\linewidth]{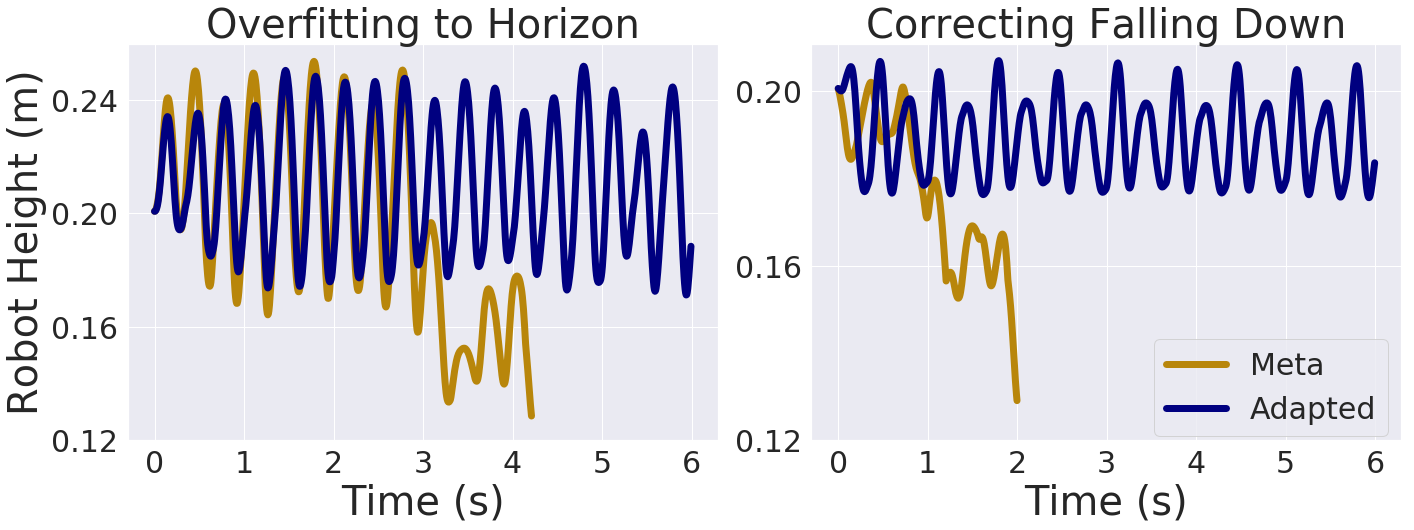}
    \caption{\label{fig:overfitting_fallingdown} Adaptation corrects falling down, due to changes in horizon length, or unstable environment dynamics.}
  \end{subfigure}
  \begin{subfigure}[t]{\linewidth}
   \centering
    \includegraphics[keepaspectratio, width=\linewidth]{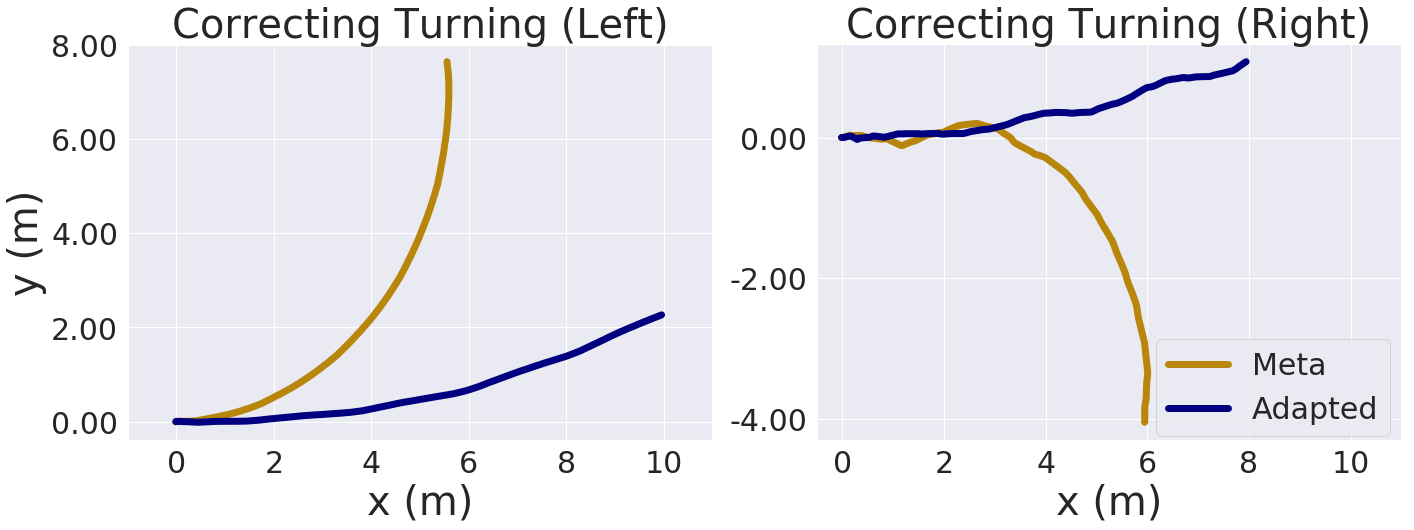}
    \caption{\label{fig:correcting_turning} Adaptation corrects turning, due to changes in environment variables such as battery voltages and motor strengths.}
  \end{subfigure}
  \caption{\label{fig:qualitative_adaptation}Examples of ES-MAML adaptation in the Minitaur simulation environment. \textbf{(a)}: HC corrects the robot from falling, leading to longer and more stable walking behavior. Falling behavior can be caused by overfitting to horizon length (e.g. stopping halfway at time=3s), or unstable dynamics. \textbf{(b)}: HC corrects the robot from curving to the left or right, by straightening the walking trajectory.}
\end{figure}
\label{subsec:qualitative}
To demonstrate that our algorithm can adapt policies to changes in dynamics, we use disjoint sets of random seeds for the training and testing environment populations, which generate environment parameters found in Table \ref{table:parameter_range}. Furthermore, we increase the episode length of testing to 1000 steps from 500 used in training because we find that training can overfit to the episode length: the robot fails soon after 500 steps (Figure~\ref{fig:overfitting_fallingdown}, Left).
\begin{figure}[t]
    \centering
    \includegraphics[keepaspectratio, width=\linewidth]{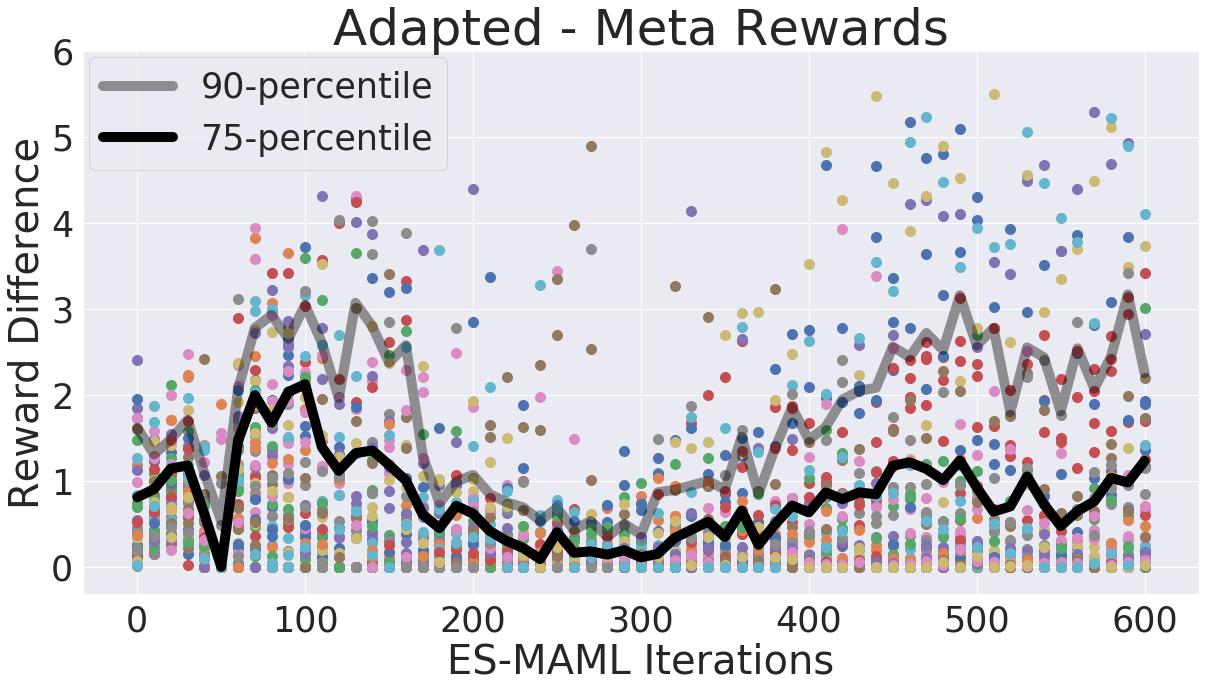}
    \caption{\label{fig:normaltrain_normaltest_scatterplot} Scatterplot showing adaptation gaps for 50 test tasks while training with ES-MAML. The same colored dots correspond to the same test task. Percentiles are plotted to display distribution of gaps.}
\label{fig:scatterplot}
\end{figure} 

In new test environments, the meta-policy can fail in various ways depending on the environment parameters. For example, the robot falls before the episode ends, or veers off to the left or right (Figure~\ref{fig:correcting_turning}). Our method adapts effectively in the new environments; by using only 50 rollouts, by keeping the robot balanced while walking straight. Figure \ref{fig:scatterplot} shows that for a large variety of new environments, as the number of ES-MAML iterations increases, the adaption becomes more effective and achieves greater performance gains.

\subsection{Comparison with PG-MAML}
We perform comparisons between our method and PG-MAML across two different adaptation tasks, with more distribution difference between training and testing (Table~\ref{tab:pg_comparison_tests}). In the \emph{Uniform Test} setting, the training and testing environments sample different physical parameters within the same range. In the \emph{Extreme Test} setting, the testing environment only samples physical parameters from the extremities of the range. As an extra baseline, we also train robust policies using Domain Randomization (DR) \cite{sim-to-real}. For a fair comparison, we use the same amount of experience (50 rollouts) for adaptation between ES-MAML  and PG-MAML, where ES-MAML uses these for Hill-Climbing, while PG-MAML performs two gradient steps.
\begin{table}[t]
    \centering
    \scalebox{1.05}{
    \begin{tabular}{|c|c|c|}
        \hline
        \textbf{Name} &\textbf{Train Task} & \textbf{ Test Task} \\\hline
        Uniform Test & Entire range & Entire range \\
        \hline
        Extreme Test & Entire range & $\{\min,\max\}$ of range  \\
         \hline
   %     Mass-Motor Train & Mass, Motor Strengths  & Foot friction,  \\
%        Friction Test &  &  Control Latencies  \\\hline
    \end{tabular}
    }
    \caption{Test setup for PG-MAML Comparison. All environment parameters were sampled uniformly.}
    \label{tab:pg_comparison_tests}
\end{table}

\begin{figure}[b]
    \centering
    \includegraphics[keepaspectratio, width=\linewidth]{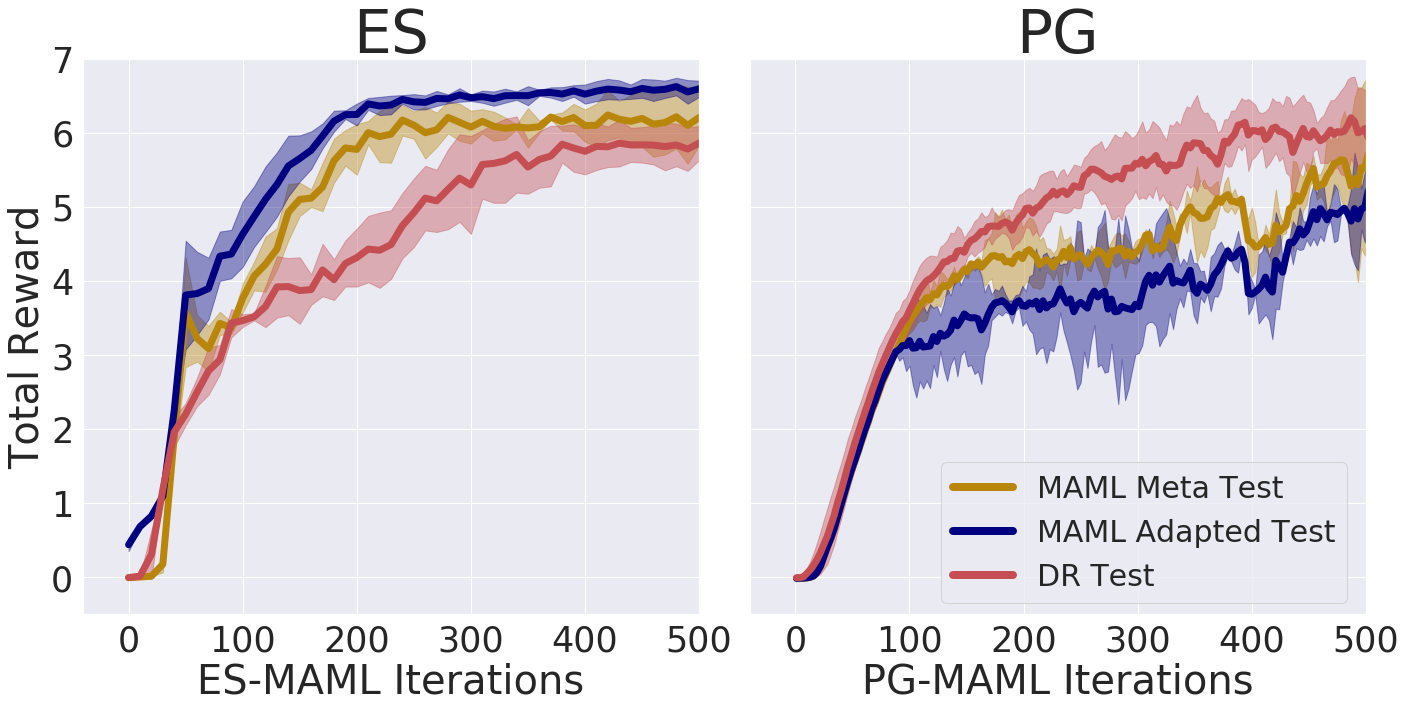}
    \caption{\label{fig:normaltrain_normaltest} \textbf{Uniform Test} for ES and PG. We see that ES-MAML's adaptation performance is the highest, outperforming its meta-policy and domain randomization with ES. On the other hand, the trend is reversed for PG-MAML, which possesses a negative adaptation gap and a very strong domain randomized policy.}
\end{figure}

\begin{figure}[t]
    \centering
    \includegraphics[keepaspectratio, width=0.48\textwidth]{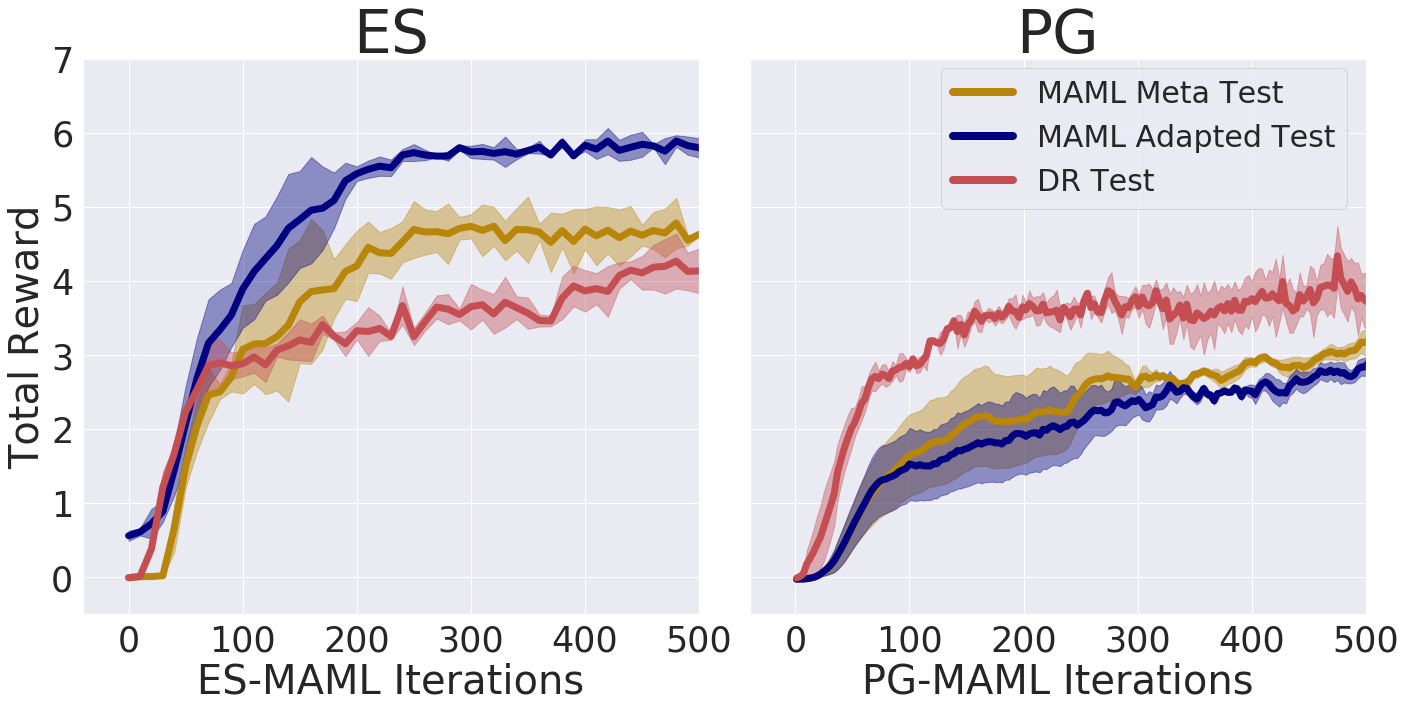}
    \caption{\textbf{Extreme Test} for ES and PG. The extreme test environment increases the adaptation gap for ES-MAML, while PG-MAML produces lower gains everywhere, with its domain randomization performance remaining the strongest.}  
    \label{fig:normaltrain_extremetest}
\end{figure}

In both adaptation tasks, our method's adapted policy improves performance: The blue curve is always above the brown curve in Figure \ref{fig:normaltrain_normaltest} and \ref{fig:normaltrain_extremetest}. 
As expected, the performance of the meta-policy decreases significantly in the Extreme Test (Fig.~\ref{fig:normaltrain_extremetest}). Even on this more difficult environment, our method still adapts significantly and brings the adapted policy to a similar performance as Uniform Test.

In contrast for PG-MAML, the adaptation fails to perform any improvement in both cases. This is likely caused by noisy gradient estimation from policy gradient methods. This observation is consistent with poor adaptation performance on the real robot in Section~\ref{subsec:real_robot}. While PG-MAML is well known for adapting to \emph{reward} changes \cite{norml}, its failure in this setting is potentially due to \emph{dynamics} changes.

\subsection{Adaptation in the Presence of Noise} \label{subsec:noise_adaptation}
\begin{figure}[ht]
  \begin{subfigure}{\linewidth}
    \centering
    \includegraphics[keepaspectratio, width=\linewidth]{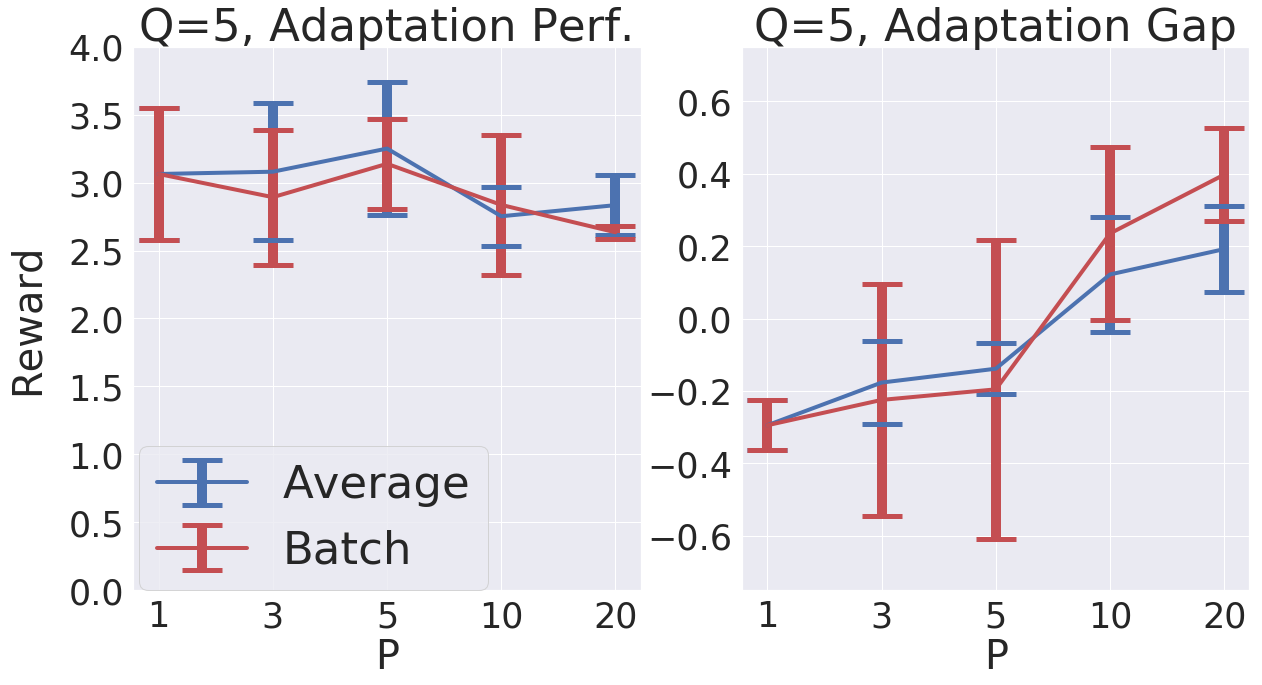}
    \caption{\label{fig:minitaur_noisy}In \textbf{Minitaur-Noisy}, Batch produces a higher adaptation gap than Average, especially when increasing $P$, while still maintaining approximately same level of adaptation performance.}
  \end{subfigure}
  \begin{subfigure}{\linewidth}
    \centering
    \includegraphics[keepaspectratio, width=\linewidth]{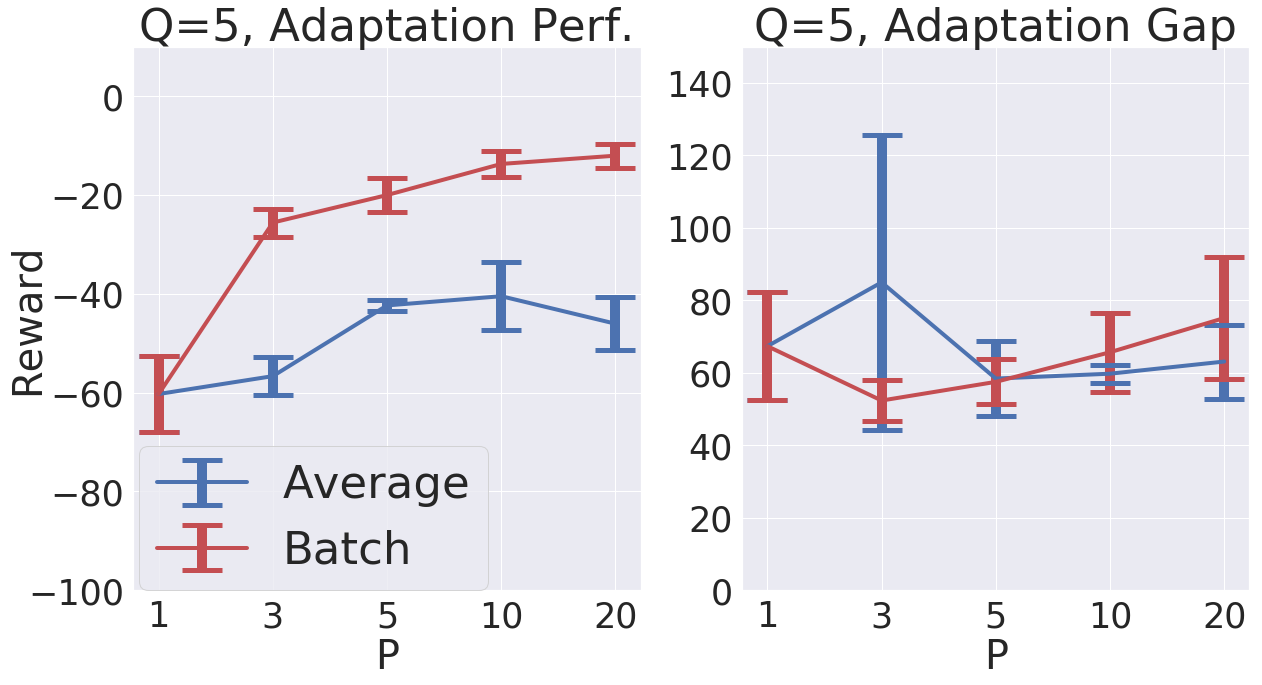}
    \caption{\label{fig:nav_2d}In \textbf{Obs-Noise Nav-2D}, Batch outperforms Averaging when measuring mean adaptation performance across entire test distribution after training convergence.}
  \end{subfigure}
  \caption{Comparison between Batch and Average Hill-Climbing on the Obs-Noise Nav-2D and Minitaur-Noisy environments. X-Axis: Parallel evaluations used during a Hill-Climbing adaptation step. Y-Axis: Total Reward.}
\end{figure}

We further compare our Hill-Climbing variants (``Average'' and ``Batch'') in the stochastic Minitaur simulator (Fig.~\ref{fig:minitaur_noisy}). In the real robot experiments, we found that the sources of noise may include different initializations, drift of dynamics over time (e.g. due to motor overheating and battery level dropping), sensor noise, and environmental perturbations (e.g. ground with nonuniform textures and friction). To mimic real world scenarios, we further increase the stochasticity of our simulation. On top of observation noise and random initialization that are mentioned in Subsection~\ref{subsec:task_setup}, we perturbed the robot with random forces so that the variance in rewards across multiple runs for the same policy is comparable to that observed on the real robot. In this highly noisy environment, we found that although the final performances are comparable, ``Batch'' Hill-Climbing generally produces more obvious adaptations than its ``Average'' counterpart. Increasing $P$ generally improved adaptation, with ``Batch'' producing larger adaptation gains than ``Average'' when $P>5$.

To demonstrate the generality of our method, we performed the same comparison in a different noisy environment. We created a noisy version of the widely-used Nav-2D environment from \cite{PG-MAML}. We added Gaussian noises (0.0 mean, 1.0 std.) to the observations while keeping all other setups unchanged. In Fig.~\ref{fig:nav_2d}, ``Batch'' Hill-Climbing significantly improves final performance after adaptation, and the performance gain scales with $P$.

\section{Conclusion}
Legged robots need to quickly adapt their locomotion skills to move in different dynamic environments. In this paper, we present an evolutionary meta-learning algorithm that enables locomotion policies to quickly adapt in noisy real world scenarios. The core idea is to develop an efficient and noise-tolerant adaptation operator, and integrate it into meta-learning frameworks. We have shown that this Batch Hill-Climbing operator works better in handling noise than simply averaging rewards over multiple runs. Our algorithm has achieved greater adaptation performance than the state-of-the-art MAML algorithms that are based on policy gradient. Finally, we validate our method on a real quadruped robot. Trained in simulation, the locomotion policies can successfully adapt to two real-world robot environments, whose dynamics have been drastically changed.

In the future, we plan to extend our method in several ways. First, we believe that we can replace the Gaussian perturbations in the evolutionary algorithm with non-isotropic samples to further improve the sample efficiency during adaptation. With less robot data required for adaptation, we plan to develop a lifelong learning system, in which the robot can continuously collect data and quickly adjust its policy to learn new skills and to operate optimally in new environments.

\ignorethis{In this work, we have demonstrated that model-free adaptation to changes in dynamics is possible in the real world, even when dealing with highly stochastic and noisy environments such as for legged robots. In comparison to other works which use domain randomization for robots to perform in the real world, we used the paradigm of Model Agnostic Meta-Learning (MAML). This required modifying ES-MAML with a noise-reductive adaptation procedure termed \textit{Batch Hill-Climbing}, which is both empirically and theoretically sound. In comparison, PG-MAML was unable to adapt to real world tasks and was shown in simulation to be stronger for domain randomization, rather than adaptation. We believe that this close relationship between PG-MAML and domain randomization is an important question, and it remains to be understood in future work. 

Furthermore, there are numerous ways to extend our method. For instance, we suspect that we can replace our Gaussian perturbations in Batch Hill-Climbing with non-isotropic samples in order to improve the sample complexity during adaptation. Since we have shown the general ES-MAML paradigm to be effective for real robotics, there is hope that ES-MAML can also be modified to an online adaptation setting, in which there are a \textit{sequence} of incoming real robotics tasks. We believe the ES-MAML paradigm is a promising new way to perform model-free meta-learning on real robotics. 
}

\clearpage
\section*{Acknowledgements}
\noindent We thank Vikas Sindhwani, Daniel Seita, and the Google AI Blog Team for fruitful discussions when writing this work.
\vspace{-0.1cm}
%% Use plainnat to work nicely with natbib. 
\renewcommand*{\bibfont}{\footnotesize}
\bibliographystyle{plainnat}
\bibliography{references}

%\clearpage
\section{Appendix}
In this section we prove our main theoretical result regarding $P$-batch Hill-Climbing method. For ease of reading, we reset all notation from the main body of this paper, and formally re-define all terms. In particular, we use $T$ to denote the iteration length of our algorithm rather than for the notation for a task, and rather than also using the \textit{query number} $Q$. Similarly, we use $\theta_{t}$ rather than $\theta^{(q)}$, and also $\sigma$ rather than the adaptation step-size $\alpha$ for scaling on Gaussians.

To begin, we need a definition of $(\mu,\rho)$-strong concavity that we give below:

%\begin{definition}[$L$-Lipschitz functions]
%We say that function $f:\mathbb{R}^{d} \times \mathbb{R}$ is %$L$-Lipschitz in the open subset $\mathcal{B} \subseteq %\mathbb{R}^{d}$ if the following holds for every %$\mathbf{x},\mathbf{y} \in \mathcal{B}$:
%\begin{equation}
%|f(\mathbf{y})-f(\mathbf{x})| \leq L\|\mathbf{y}-\mathbf{x}\|_{2} %\end{equation}
%\end{definition}

%\begin{definition}[$\lambda$-smoothness]
%We say that differentiable function $f:\mathbb{R}^{d} \rightarrow %\mathbb{R}$ is $\lambda$-smooth in the open subset $\mathcal{B} %\subseteq \mathbb{R}^{d}$ if the following holds for every %$\mathbf{x},\mathbf{y} \in \mathcal{B}$:
%\begin{equation}
%\|\nabla f(\mathbf{y}) - \nabla f(\mathbf{x})\|_{2} \leq \lambda %\|\mathbf{y}-\mathbf{x}\|_{2}.    
%\end{equation}
%\end{definition}

\begin{definition}[$(\mu,\rho)$-strong concavity]
We say that differentiable concave function $f:\mathbb{R}^{d} \rightarrow \mathbb{R}$ is $(\mu,\rho)$-strongly concave for some $0 < \mu < \rho$ in the open subset $\mathcal{B} \subseteq \mathbb{R}^{d}$ if the following holds for every $\mathbf{x},\mathbf{y} \in \mathcal{B}$:
\begin{equation}
f(\mathbf{y}) \leq f(\mathbf{x}) + \nabla f(\mathbf{x})^{\top}(\mathbf{y}-\mathbf{x}) - \frac{\mu}{2}\|\mathbf{y}-\mathbf{x}\|^{2}    
\end{equation}
and
\begin{equation}
f(\mathbf{y}) \geq f(\mathbf{x}) + \nabla f(\mathbf{x})^{\top}(\mathbf{y}-\mathbf{x}) - \frac{\rho}{2}\|\mathbf{y}-\mathbf{x}\|^{2}    
\end{equation}
\end{definition}

Below we restate our result for the reader's convenience.
We assume that sampled Gaussian directions $\mathbf{g}^{i}$ are normalized so that they all have norm $\sqrt{d}$ (in the unnormalized case the expected squared norm is $d$). Analogous result can be proven in the unnormalized setting, yet the proof is more technical.

\begin{theorem}
\label{hill}
Assume that $f:\mathbb{R}^{d} \rightarrow \mathbb{R}$ is a 
$(\mu,\rho)$-strong concave function in compact domain $\Omega \subseteq \mathbb{R}^{d}$ of interest for some $0 < \mu < \rho$ and that $\theta \rightarrow \nabla f(\theta)$ is continuous in $\Omega$.
Assume furthermore that $\mathrm{diam}(\Omega) \leq D$
and that our algorithm, for an input $\theta$, observes noisy objective $\widetilde{f}(\theta, \varepsilon) = f(\theta) + \varepsilon$, where $\varepsilon$ is an external noise variable from the environment, which can be adversarial. Take Hill-Climbing Algorithm consisting of $T$ steps and with parallel batch $P$ evaluations on $\widetilde{f}$. Assume that in every iteration, at most $W$ ($W<P$) perturbations are characterized by unbounded negative error $\varepsilon$ and for remaining $P-W$ we have: $|\varepsilon| \leq \Lambda$.
Assume that $\sup_{\theta \in \Omega} \|\nabla f(\theta)\|_{2} \leq L$ for some $L > 0$. Take $\phi>0$ such that: $\phi > \frac{4(\rho-\mu)\sqrt{Td}\sigma}{7}+\frac{16 \Lambda \sqrt{T}}{7\sigma \sqrt{d}}$.
Denote by $\theta^{\mathrm{opt}}$ the optimal solution (maximizing $f(\theta)$) and by $\theta_{0},\theta_{1},...$ a sequence of points found by the algorithm. 
Then for any $s>0$, if $P \geq W + e^{3sd\log(d)\sqrt{T}}$ and
$\sigma \leq 4\sqrt{\frac{\Lambda}{7\sqrt{d}}}$, the following holds with probability $p \geq 1 - Te^{-s}$:
\begin{align}
\begin{split}
\frac{\sum_{t=0}^{T-1}(f(\theta^{\mathrm{opt}})-f(\theta_t))}{T} \leq \\
\frac{1}{\sqrt{T}}(\frac{D^{2}}{2}+(L+\frac{4L^{2}}{\sqrt{T}})^{2}+8DL^{2})+D \phi
\end{split}
\end{align}
or $|f(\theta^{\mathrm{opt}}) - f(\theta_{i})| \leq D \phi$
for some $\theta_{i}$.

\end{theorem}

We see that if $\sigma \gg \Lambda$, then the algorithm converges to or visits the solution close to optimal $\theta^{*}$ (in terms of the value of $f$) for small enough $\Lambda$.

\begin{remark}
Note that in practice one can reduce the number of samples $P$ in the batch needed by performing non-isotropic sampling, where next samples are obtained with higher probabilities from these regions, where previous samples led to high rewards. In practice we observed that such a mechanism is not needed and good quality solution can be still found with relatively small batch-sizes.
\end{remark}

Below we prove Theorem \ref{hill}.

\begin{proof}
Denote: $\mathcal{T}_{\phi}=\{\theta \in \Omega: \|\nabla f(\theta)\|_{2} \leq \phi\} \subseteq \Omega$. Note that $\mathcal{T}_{\phi}$ is compact.
If $\theta_{i} \in \mathcal{T}_{\phi}$ for some $i$ then, since $\mathrm{diam(\Omega)} \leq D$, we get:
$|f(\theta^{*}) - f(\theta_{i})| \leq D \phi$ and thus we are done. Denote: $\mathcal{W}_{\phi}=\{\theta \in \Omega: \|\nabla f(\theta)\|_{2} \geq \phi\} \subseteq \Omega$.
Fix a particular step $t$ of the Hill-Climbing procedure. 
Denote by $\mathbf{g}_{t}^{*}$ the Gaussian vector associated with the perturbation $\sigma \mathbf{g}_{t}^{*}$ that was chosen during Hill-Climbing in that particular step. Denote by $\{\mathbf{g}_{t}^{1},...,\mathbf{g}_{t}^{P}\}$ the set of all random vectors used to create all perturbations in that step, by $\mathcal{S}$ its subset consisting of those of them for which the corresponding noisy evaluations are within distance $\Lambda$ from the exact value and by $\theta_t \in \mathbb{R}^{d}$ vectorized policy in step $t$.
By the definition of $\mathbf{g}^{*}_{t}$ we have:
\begin{equation}
\label{max_eq}
f(\theta_{t} + \sigma \mathbf{g}_{t}^{*}) \geq \max_{i \in \{1,...,P\}} f(\theta_{t} + \sigma \mathbf{g}_{t}^{i}) - 2\Lambda
\end{equation}

Denote by $\mathcal{A}^{i,t}_{\phi}$ an event that $\text{angle} (\mathbf{g}^{i}_{t},\mathbf{v}_{t}) \leq \phi$, where $\mathbf{v}_{t}= \nabla f(\theta_{t})$ and $\text{angle} (\mathbf{x},\mathbf{y})$ stands between an angle between $\mathbf{x}$ and $\mathbf{y}$.
Note that since each $\mathbf{g}^{i}_{t}$ is sampled from isotropic probabilistic distribution on $\mathbb{R}^{d}$, we conclude that:
\begin{equation}
\cos(\text{angle} (\mathbf{g}^{i}_{t},\mathbf{v}_{t})) \sim \frac{g(1)}{\sqrt{\sum_{i=1}^{d}g(i)^{2}}}
\end{equation}
where $g(1),...,g(d) \sim \mathcal{N}(0,1)$.
Furthermore, random variables $\cos(\text{angle} (\mathbf{g}^{i}_{t},\mathbf{v}_{t}))$ for $i=1,2,...,$, $t=1,2,...$ are independent (since perturbations are chosen independently).

Denote by $\mathcal{A}_{\phi}^{t}$ an event that $\mathcal{A}^{i,t}_{\phi}$ holds for \textbf{at least one} $i$ such that $\mathbf{g}_{t}^{i} \in \mathcal{S}$ (e.g. the corresponding noisy evaluation is within distance $\Lambda$ from the exact value). 

Take some $0 < \phi_{0} < \frac{\pi}{4}$.
Denote by $\mathcal{B}^{t}_{\phi_{0}}$ an event that $\text{angle} (\mathbf{g}^{*}_{t}, \mathbf{v}_{t}) > 2 \phi_{0}$.

Our first goal is to understand what upper bound on the length of the gradient $\|\mathbf{v}_{t}\|_{2}$ is implied by the condition that  $\mathcal{B}^{t}_{\phi_{0}} \cap \mathcal{A}^{t}_{\phi_{0}}$ holds.

So assume that $\mathbb{P}[\mathcal{B}^{t}_{\phi_{0}} \cap \mathcal{A}^{t}_{\phi_{0}}]>0$.
Note that if $\mathcal{A}^{t}_{\phi_{0}}$ holds then one of the events $\mathcal{A}^{i,t}_{\phi_{0}}$ holds for such $i$ that the corresponding noisy evaluation is within distance $\Lambda$ from the exact value. Take that $i$ and call it $i^{*}$. From $(\mu,\rho)$-strong concavity, by the definition of $i^{*}$ and from the fact that $\mathcal{B}^{t}_{\phi_{0}}$ holds, we have:
\begin{equation}
f(\theta_t + \sigma \mathbf{g}^{*}_{t}) \leq f(\theta_t) + \|\mathbf{v}_{t}\|_{2}\sigma \|\mathbf{g}^{*}\|_{2}\cos(\beta_1) - \frac{\mu}{2} \sigma^{2}  \|\mathbf{g}^{*}\|_{2}^{2}
\end{equation}
and
\begin{equation}
f(\theta_t + \sigma \mathbf{g}_{t}^{i^{*}}) \geq f(\theta_t) + \|\mathbf{v}_{t}\|_{2}\sigma \|\mathbf{g}^{i^{*}}\|_{2}\cos(\beta_2) - \frac{\rho}{2}\sigma^{2}  \|\mathbf{g}^{i^{*}}\|_{2}^{2}  
\end{equation}
for some $0 < \beta_1, \beta_2 < \frac{\pi}{2}$ such that $\beta_1 > 2\phi_{0}$ and $\beta_2 < \phi_{0}$.
Thus, by Eq. \ref{max_eq}, we have:
\begin{align}
\begin{split}
\|\mathbf{v}_{t}\|_{2}\sigma \|\mathbf{g}^{*}\|_{2}\cos(\beta_1) - \frac{\mu}{2} \sigma^{2}  \|\mathbf{g}^{*}\|_{2}^{2} \geq \\ \|\mathbf{v}_{t}\|_{2}\sigma \|\mathbf{g}^{i^{*}}\|_{2}\cos(\beta_2) - \frac{\rho}{2}\sigma^{2}  \|\mathbf{g}^{i^{*}}\|_{2}^{2} + 2\Lambda 
\end{split}
\end{align}
Therefore we get:
\begin{align}
\begin{split}
\sigma\|\mathbf{v}_{t}\|_{2}( \|\mathbf{g}^{i^{*}}\|_{2}\cos(\beta_2)-\|\mathbf{g}^{*}\|_{2}\cos(\beta_1)) \leq \\ \frac{\rho}{2}\sigma^{2}  \|\mathbf{g}^{i^{*}}\|_{2}^{2} - \frac{\mu}{2} \sigma^{2}  \|\mathbf{g}^{*}\|_{2}^{2} + 2\Lambda
\end{split}
\end{align}
Now we can use the fact that vectors $\mathbf{g}^{i}_{t}$ are normalized, and we get:
\begin{equation}
\|\mathbf{v}_{t}\|_{2} \leq \frac{\rho-\mu}{2\tau}\sqrt{d}\sigma + \frac{2\Lambda}{\sigma \tau \sqrt{d}}
\end{equation}
where $\tau = \cos(\phi_{0}) - \cos(2\phi_{0})$.
To summarize, we conclude that:
\begin{remark}
\label{imp_remark}
If $\mathcal{B}^{t}_{\phi_{0}} \cap \mathcal{A}^{t}_{\phi_{0}}$ holds then:
\begin{equation}
\|\mathbf{v}_{t}\|_{2} \leq \frac{\rho-\mu}{2\tau}\sqrt{d}\sigma + \frac{2\Lambda}{\sigma \tau \sqrt{d}} 
\end{equation}
where $\tau = \cos(\phi_{0}) - \cos(2\phi_{0})$.
\end{remark}
Denote by $\theta^{*}$ an optimal solution in $\mathcal{W}_{\phi}$.
Fix a sequence of positive numbers $\{\eta_t\}_{t=0,1,...,}$
Now let us rewrite $\sigma \mathbf{g}^{i^{*}}_{t}$ as:
$\sigma \mathbf{g}^{i^{*}}_{t} = \eta_t \widehat{\mathbf{g}}$, where $\widehat{\mathbf{g}}_{t}=\frac{\sigma \mathbf{g}^{i^{*}}_{t}}{\eta_t}$.
Note that we have:
\begin{align}
\begin{split}
\label{cos_thm}
\|\theta_{t+1} - \theta^{*}\|^{2} = \|\theta_t + \sigma \mathbf{g}^{i^{*}}_{t} - \theta^{*}\|_{2} = \\
\|\theta_t - \theta^{*}\|_{2}^{2} + \sigma^{2}\|\mathbf{g}^{i^{*}}_{t}\|^{2}_{2} - 2\sigma \mathbf{g}^{i^{*}}_{t}(\theta^{*}-\theta_t) = \\
\|\theta_t - \theta^{*}\|_{2}^{2} + \eta_{t}^{2}\|\widehat{\mathbf{g}}_{t}\|^{2}_{2} - 2\eta_{t} \widehat{\mathbf{g}}_{t}(\theta^{*}-\theta_t)
\end{split}
\end{align}
Denote by $A_t>0$ some upper bound on $\|\widehat{\mathbf{g}}_{t}-\mathbf{v}_{t}\|_{2}$.
We thus have:
\begin{equation}
\|\widehat{\mathbf{g}}_{t}-\mathbf{v}_{t}\|_{2} \leq A_t    
\end{equation}
From the triangle inequality we have:
\begin{equation}
\label{tri}
\|\widehat{\mathbf{g}}_{t}\|_{2}^{2} \leq (\|\mathbf{v}\|_{2} + A_t)^{2}
\end{equation}
Note that from Cauchy-Schwartz inequality, we have:
\begin{equation}
\label{sup_eq}
\sup_{\theta \in \Omega}|(\theta - \theta_t)^{\top}\widehat{\mathbf{g}}_{t} - (\theta - \theta_t)^{\top}\mathbf{v}_{t}| \leq DA_t  
\end{equation}
where $D$ is a diameter of $\Omega$.
From Equality \ref{cos_thm} we get:
\begin{equation}
2\eta_{t} \widehat{\mathbf{g}}_{t}(\theta^{*}-\theta_t) = 
\|\theta_t - \theta^{*}\|_{2}^{2} - \|\theta_{t+1} - \theta^{*}\|^{2}  +
\eta_{t}^{2}\|\widehat{\mathbf{g}}_{t}\|^{2}_{2}.
\end{equation}
Therefore, from the above and by Eq. \ref{tri}, we get:
\begin{equation}
2 \widehat{\mathbf{g}}_{t}(\theta^{*}-\theta_t) \leq 
\frac{\|\theta_t - \theta^{*}\|_{2}^{2} - \|\theta_{t+1} - \theta^{*}\|^{2}}{\eta_t} + \eta_t (\|\mathbf{v}\|_{2} + A_t)^{2}    
\end{equation}
Therefore, by Eq. \ref{sup_eq}, we get:
\begin{align}
\begin{split}    
2\mathbf{v}_{t}^{\top}(\theta^{*} - \theta_t) \leq     
\frac{\|\theta_t - \theta^{*}\|_{2}^{2} - \|\theta_{t+1} - \theta^{*}\|^{2}}{\eta_t} + \\ \eta_t (\|\mathbf{v}\|_{2} + A_t)^{2} + 2DA_{t}
\end{split}
\end{align}
From concavity of $f$, we obtain:
\begin{equation}
f(\theta^{*}) - f(\theta_t) \leq \mathbf{v}_{t}^{\top}(\theta^{*}-\theta_t)    
\end{equation}
Combined together, the last two inequalities give us:
\begin{align}
\begin{split}
2(f(\theta^{*})-f(\theta_t)) \leq     
\frac{\|\theta_t - \theta^{*}\|_{2}^{2} - \|\theta_{t+1} - \theta^{*}\|^{2}}{\eta_t} + \\ \eta_t (\|\mathbf{v}\|_{2} + A_t)^{2} + 2DA_{t}
\end{split}
\end{align}
Therefore we obtain:
\begin{align}
\begin{split}
2\sum_{t=0}^{T-1}(f(\theta^{*})-f(\theta_t)) \leq
\sum_{t=0}^{T-1}
\frac{\|\theta_t - \theta^{*}\|_{2}^{2} - \|\theta_{t+1} - \theta^{*}\|^{2}}{\eta_t} 
+ \\ \sum_{t=0}^{T-1}(\eta_t (\|\mathbf{v}\|_{2} + A_t)^{2} + 2DA_{t}) = \\
\sum_{t=0}^{T-1}\|\theta_t-\theta^{*}\|_{2}^{2}(\frac{1}{\eta_t} - \frac{1}{\eta_{t-1}}) + \sum_{t=0}^{T-1}(\eta_{t}(L+A_{t})^{2}+2DA_{t})
\end{split}
\end{align}
where $L$ stands for the upper bound on $\|\nabla f(\theta)\|_{2}$ in the domain $\Omega$ and we take $\eta_{-1} = \infty$.

We are ready to complete the proof. 
Now denote: $\mathcal{B}_{\phi_{0}} = \bigcup_{t=1}^{T} \mathcal{B}^{t}_{\phi_{0}}$
and take $\mathcal{G}_{\phi_{0}}$ to be the complement of $\mathcal{B}_{\phi_{0}}$.
By the union bound, we get:
\begin{equation}
\label{union_bound}
\mathbb{P}[\mathcal{G}_{\phi_{0}}] \geq 1 - \sum_{t=1}^{T}\mathbb{P}[\mathcal{B}^{t}_{\phi_{0}}]
\end{equation}

Let us assume now that $\mathcal{G}_{\phi_{0}}$ holds.
From our analysis so far we can conclude that with probability $p^{*} \geq 1-\sum_{t=1}^{T}\mathbb{P}[\mathcal{B}^{t}_{\phi_{0}}]$ the following event $\mathcal{E}$ holds:
\begin{align}
\begin{split}
\label{aveg}
\frac{\sum_{t=0}^{T-1}(f(\theta^{*})-f(\theta_t))}{T} \leq
\frac{1}{2T}\sum_{t=0}^{T-1}\|\theta_t-\theta^{*}\|_{2}^{2}(\frac{1}{\eta_t} - \frac{1}{\eta_{t-1}}) + \\ \frac{1}{2T}\sum_{t=0}^{T-1}(\eta_{t}(L+A_{t})^{2}+2DA_{t}) \end{split}
\end{align}
Notice that:
\begin{equation}
p \geq 1 - \sum_{t=1}^{T}(\mathbb{P}[\mathcal{B}^{t}_{\phi_{0}} \cap \mathcal{A}^{t}_{\phi_{0}}] + \mathbb{P}[\mathcal{B}^{t}_{\phi_{0}} \cap \lnot \mathcal{A}^{t}_{\phi_{0}}])
\end{equation}
where $\lnot \mathcal{H}$ stands for the complement of an event $\mathcal{H}$.
Now, if $\phi$ satisfies: 
\begin{equation}
\label{l_ineq}
\phi > \frac{\rho - \mu}{2\tau}\sqrt{d}\sigma + \frac{2\Lambda}{\sigma \tau \sqrt{d}},    
\end{equation}
then by Remark \ref{imp_remark}, we have:
$\mathbb{P}[\mathcal{B}^{t}_{\phi_{0}} \cap \mathcal{A}^{t}_{\phi_{0}} \cap \{\theta_{t} \in \mathcal{W}_{\phi}\}] = 0$ for every $t$.
Therefore we can conclude that if Inequality \ref{l_ineq} holds then with probability:
\begin{equation}
p \geq 1 - \sum_{t=1}^{T}\mathbb{P}[\mathcal{B}^{t}_{\phi_{0}} \cap \lnot \mathcal{A}^{t}_{\phi_{0}}] \geq 
1 - \sum_{t=1}^{T}\mathbb{P}[\lnot \mathcal{A}^{t}_{\phi_{0}}]    
\end{equation}
event $\mathcal{E} \cup \mathcal{F}$ holds, where $\mathcal{F}$ is an event that for some $i$ we have: $|f(\theta^{\mathrm{opt}}) - f(\theta_{i})| \leq D \phi$. 

Now let us take an event $\lnot \mathcal{A}^{t}_{\phi_{0}}$.
By the definition of $\mathcal{A}^{t}_{\phi_{0}}$, we have:
\begin{equation}
\mathbb{P}[\lnot \mathcal{A}^{t}_{\phi_{0}}] =
\prod_{i=1}^{P - W}\mathbb{P}[\lnot \mathcal{A}^{i,t}_{\phi_{0}}]
\end{equation}
where $W$ stands for the number of those perturbations for which noisy evaluation is not within distance $\Lambda$ from the exact value (without loss of generality we assume that these are the last $W$ perturbations).

Using the well-known inequalities estimating tail of Gaussian distribution: 
$(\frac{1}{x}-\frac{1}{x^{3}})\frac{e^{-\frac{x^{2}}{2}}}{\sqrt{2\pi}} \leq \mathbb{P}[g > x] \leq \frac{1}{x}\frac{e^{-\frac{x^{2}}{2}}}{\sqrt{2\pi}}$, where $g \sim \mathcal{N}(0,1)$ we conclude that for every $x >0$ 
and $i \leq P-W$ 
we have:
\begin{equation}
\mathbb{P}[\mathcal{A}^{i,t}_{\phi_{0}}] \geq   
2(\frac{1}{s}-\frac{1}{s^{3}})\frac{e^{-\frac{s^{2}}{2}}}{\sqrt{2 \pi}}(1-\frac{2}{x}\frac{e^{-\frac{x^{2}}{2}}}{\sqrt{2 \pi}}d)
\end{equation}
where $s=\sqrt{\frac{d}{1-\cos^{2}(\phi_{0})}}x$.

We conclude that for every $x>0$, if $\phi > \frac{\rho - \mu}{2\tau}\sqrt{d}\sigma + \frac{2\Lambda}{\sigma \tau \sqrt{d}}$, where $\tau=\cos(\phi_{0})-\cos(2 \phi_{0})$, then
with probability $p \geq 1 - T(1-2(\frac{1}{s}-\frac{1}{s^{3}})\frac{e^{-\frac{s^{2}}{2}}}{\sqrt{2 \pi}}(1-\frac{2}{x}\frac{e^{-\frac{x^{2}}{2}}}{\sqrt{2 \pi}}d))^{P-W}$, Inequality \ref{aveg} holds, where $A_{t}$ is such that:  $\|\widehat{\mathbf{g}}_{t} - \mathbf{v}_{t}\|_{2} \leq A_t$ or event $\mathcal{F}$ holds (or both).

Now, if we take: $\eta_{t} = 
\frac{\|\sigma \mathbf{g}_{t}^{i^{*}}\|_{2}}{\|\mathbf{v}_{t}\|_{2}}$, then (from the Law of Cosines), we can take any $A_{t}$ such that:
\begin{equation}
A_{t} \geq 2\|\mathbf{v}_{t}\|_{2}^{2}(1-\cos(2\phi_{0}))  
\end{equation}
Since $\|\mathbf{v}_{t}\|_{2} \leq L$ for $t=0,1,...$ and 
$1 - \cos(\theta) \leq \frac{\theta^{2}}{2}$ (from Taylor expansion), we conclude that we can take: 
\begin{equation}
A_{t} = 4L^{2}\phi_{0}^{2}    
\end{equation}
Note that, by our definition of $\eta_t$, if $\sigma$ satisfies the inequality from the statement of the theorem and $\|\mathbf{v}_{t}\|_{2} > \phi$, then $\eta_t \leq \frac{1}{\sqrt{T}}$
(notice that $\|\mathbf{g}^{i^{*}_{t}}\|_{2}=\sqrt{d}$).

Take $\phi_{0} = \frac{1}{T^{\frac{1}{4}}}$.
Notice that since $\tau = \cos(\phi_{0}) - \cos(2\phi_{0})$, from Taylor expansion we get:
\begin{equation}
\tau \geq \frac{3}{2}\phi_{0}^{2}-\frac{5}{8}\phi_{0}^{4}=
\frac{3}{2}\frac{1}{\sqrt{T}} - \frac{5}{8}\frac{1}{T}
\end{equation}

We conclude that with probability
$p \geq 1 - T(1-2(\frac{1}{s}-\frac{1}{s^{3}})\frac{e^{-\frac{s^{2}}{2}}}{\sqrt{2 \pi}}(1-\frac{2}{x}\frac{e^{-\frac{x^{2}}{2}}}{\sqrt{2 \pi}}d))^{P-W}$ we have: $\mathcal{F}$ holds or the following is true:
\begin{equation}
\frac{\sum_{t=0}^{T-1}(f(\theta^{*})-f(\theta_t))}{T} \leq 
\frac{1}{\sqrt{T}}(\frac{D^{2}}{2}+(L+\frac{4L^{2}}{\sqrt{T}})^{2}+8DL^{2})
\end{equation}
or both (as long as inequalities from the statement of the theorem are satisfied).

Finally, notice that we clearly have: $\nabla f(\theta^{*}) = \phi$ and therefore: $\|f(\theta^{*})-f(\theta^{\mathrm{opt}})\| \leq D\phi$. The equality comes from the fact that $f$ is concave, $\mathcal{W}_{\phi}$ is compact and $\nabla f$ is continuous.

Now, we can take $x=\sqrt{2 \log(d)}$ and we get: $s \leq \sqrt{\frac{48 d \log(d)\sqrt{T}}{11}}$. Consequently, we have:
$p \geq 1 - Te^{-s}$ for $P-W \geq e^{3s d \log(d) \sqrt{T}}$.
The last inequality follows from the fact that function $h(x) = (1-\frac{1}{x})^{x}$ is increasing for $x>0$ and with limit (in infinity) $e^{-1}$.
That completes the proof.

\end{proof}

\end{document}